\documentclass{article}
\usepackage[preprint]{neurips_2026}
\usepackage[utf8]{inputenc}
\usepackage[T1]{fontenc}
\usepackage{amsmath,amssymb,mathtools,booktabs,multirow,graphicx,microtype}
\usepackage{algorithm,algpseudocode}
\usepackage{tikz}
\usetikzlibrary{arrows.meta,positioning,fit}
\usepackage[colorlinks=true,citecolor=blue,linkcolor=blue,urlcolor=blue]{hyperref}
\graphicspath{{figures/}}
\definecolor{agentblue}{HTML}{147D92}
\definecolor{agentorange}{HTML}{C85B39}
\definecolor{ink}{HTML}{183044}
\newcommand{\D}{\mathcal{D}}
\newcommand{\U}{\mathcal{U}}
\newcommand{\low}{\mathrm{low}}
\newcommand{\rand}{\mathrm{rand}}

\newcommand{\ind}{\mathbb{1}}
\newcommand{\smallpara}[1]{\paragraph{#1}}
\newcommand{\SelectionRows}{%
Task matched & API & 27.930 & 10.156 & -17.773 & 1/0/15 \\%
Task matched & GENERIC & 12.695 & 4.883 & -7.812 & 5/0/11 \\%
Task + action & API & 31.706 & 26.888 & -4.818 & 8/1/15 \\%
Task + action & GENERIC & 6.771 & 3.776 & -2.995 & 3/6/15 \\%
}

\newcommand{\InitializationRows}{%
101 & -5.859 & -1.562 \\%
202 & -10.156 & -3.125 \\%
303 & -2.344 & -2.734 \\%
404 & +0.000 & -4.297 \\%
505 & -1.562 & -1.953 \\%
606 & -8.984 & -4.297 \\%
}

\title{From Acting to Learning: Experience Selection\\toward Continually Learning Agents}
\hypersetup{pdftitle={From Acting to Learning: Experience Selection toward Continually Learning Agents}}
\author{
{\bfseries Yaoyuan Yan$^{1,*,\dagger}$\quad Haoxiang Jie$^{1,*}$\quad Zhiyou Heng$^{1}$\quad Kailin Wang$^{1}$}\\[2pt]
{\bfseries Yumiao Li$^{1}$\quad Lifeng Liu$^{2}$\quad Anwan Yan$^{2}$}\\[2pt]
{\normalfont $^{1}$ Country Garden Services AI Lab\quad $^{2}$ Omni AI}
}
\hypersetup{pdfauthor={Yaoyuan Yan; Haoxiang Jie; Zhiyou Heng; Kailin Wang; Yumiao Li; Lifeng Liu; Anwan Yan}}
\begin{document}
\maketitle
\begingroup
\renewcommand{\thefootnote}{\ifnum\value{footnote}=1\ensuremath{*}\else\ensuremath{\dagger}\fi}
\footnotetext[1]{Equal contribution (joint first authors).}
\footnotetext[2]{Corresponding author: \href{mailto:yanyaoyuan@bgyfw.com}{yanyaoyuan@bgyfw.com}}
\endgroup
\begin{abstract}
We study how selection rules and coverage constraints affect what a 4B language agent learns from successful interactions. The comparison covers nine selectors, two fixed experience pools, eight initialization seeds, and two sampling seeds, producing 288 trained students. Each student is evaluated on 128 development task groups with the same executor. With equal global selection freedom, 256 distinct contents, and 768 training presentations, optimizer-aware gradient selection reaches 81.45\% and 73.19\% success in the two pools. It improves on global random selection by 8.25 and 5.96 percentage points, and on global high-NLL selection by 4.10 and 4.05 points. All eight initialization-level contrasts are positive against both baselines in each pool. Relaxing task/action quotas improves gradient selection but lowers random-selection performance: the benefit of selection freedom depends on the ranking rule. A separately frozen replication on 128 reserved task groups is complete; gradient selection gains +7.03 points over random and +5.03 over high NLL after averaging the two pools equally.
\end{abstract}

\section{Introduction}
Language agents can learn from their own interaction records, but a limited training budget makes the choice of records consequential. Successful actions may repeat skills the student already has; less predictable actions may supply missing experience or introduce errors. A collector's task-success rate alone does not resolve which of its records the student should learn from.

We examine this choice using two existing experience pools, a fixed 4B student, and a common SFT procedure. The selectors rank successful action records by the student's action likelihood or by an optimizer-aware gradient score. The score follows LESS~\citep{xia2024less}. We evaluate the selected records through the student's subsequent task performance with the collection-time assistance removed.

Selection also determines coverage. A selector that can change the task and action mixture has more freedom than one constrained to fixed quotas. We compare both settings: quota-matched rules preserve a reference task/action schedule, whereas global rules can choose their own coverage. Within the global comparison, random, high-NLL, and gradient selection share the same distinct-content and presentation budgets.

Across nine selectors and 288 students, global gradient selection performs best in both pools, and its advantage persists on reserved tasks. Relaxing quotas helps gradient selection but lowers random-selection performance. The study tests experience selection within one SFT stage toward continually learning agents.

\begin{figure*}[t]
\centering
\includegraphics[width=\textwidth]{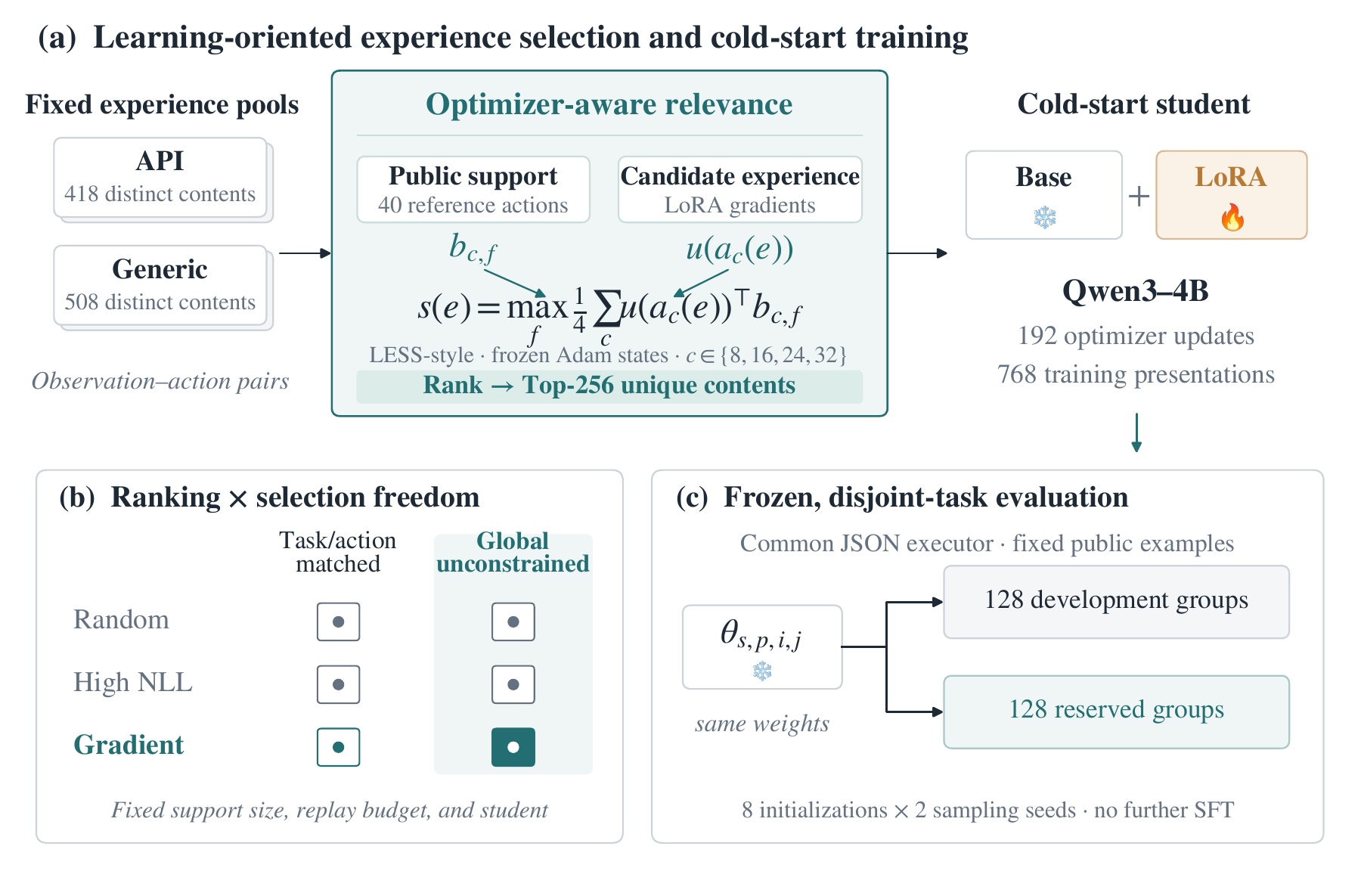}
\caption{\textbf{Experience selection and evaluation.} \textbf{(a)} A LESS-style score measures relevance to 40 public actions; selected records train a fresh student from the original base. \textbf{(b)} The condition matrix crosses ranking rules with task/action quotas or global selection. Global baselines have the same selection freedom as the gradient rule. \textbf{(c)} The resulting checkpoints are evaluated without further SFT on disjoint development and reserved panels using a common executor. Query answers are excluded from scoring. Snowflakes and flames denote frozen and trainable parameters. Collection and feature-preparation costs are reported separately.}
\label{fig:architecture}
\end{figure*}

\section{Related Work}
\label{sec:related}
\smallpara{Language agents and learning from experience.}
ALFWorld couples text interaction with household tasks~\citep{shridhar2021alfworld}, while WebShop tests grounded decisions in a shopping environment~\citep{yao2022webshop}. ReAct interleaves reasoning and actions using public demonstrations~\citep{yao2023react}. SkillScaffold estimates policy-relative skill utility and selectively injects skills during agentic reinforcement learning~\citep{zhong2026skillscaffold}. We study the selection of existing action records for supervised fine-tuning, with the deployment executor held fixed.

\smallpara{Predictability and gradient-based selection.}
SmartAD selects the lowest-student-NLL trajectory among correct teacher samples and weights action and decision segments more heavily during distillation~\citep{tang2026smartad}. Rho-1 selects training tokens using a reference model~\citep{lin2024rho}. LESS estimates targeted data influence using optimizer-aware gradient features and target demonstrations~\citep{xia2024less}. Our gradient score follows the LESS construction. We compare it with action-NLL and random selection under both matched and unrestricted coverage. Our low-NLL comparator uses action records with ordinary supervised loss. It therefore differs from SmartAD in both selection unit and loss weighting.

\smallpara{Runtime repair and harness improvement.}
Onto-EV-WM represents failed task predicates and checks proposed corrections against task conditions~\citep{wang2026ontology}. Harness Robotic OS combines robot skills, shared context, memory, and mission-level coordination for quadruped inspection~\citep{yan2026harness}. RegenHarness uses execution evidence and regression checks to accept revisions to harness configuration, including context rules, routing, and recovery policies~\citep{wang2026regenharness}. Its revision loop does not update model weights online. These systems revise runtime behavior. We instead measure how recorded text-interaction experience changes a student through SFT.

\smallpara{Selecting experience and accepting updates.}
EvoHeir separates task admission, parameter inheritance, and experience inheritance; a selected task can still lead to an unhelpful update~\citep{yang2026evoheir}. We hold training fixed and compare the records it consumes. Evaluation is sensitive to both task and training variation~\citep{agarwal2021precipice}. We report every crossed-seed condition, average sampling replicates within initialization, and evaluate the global rules on a separate reserved task panel. The reported effects are descriptive comparisons on those panels.

\section{Problem and Selection Procedure}
Let $D_s$ be a fixed historical pool, $S_p(D_s;j)$ the schedule produced by rule $p$ with sampling seed $j$, and $A_B$ the common training procedure at presentation budget $B$. Initialization/dropout seed $i$ determines the training starting state and dropout stream:
\begin{equation}
\theta_{s,p,i,j}=A_B\bigl(\theta_i,S_p(D_s;j)\bigr).
\end{equation}
With the common deployment harness $H$ and fixed panel $Q$, the behavioral endpoint is
\begin{equation}
R_{s,p,i,j}=\frac{1}{|Q|}\sum_{q\in Q}\mathbf{1}\{\operatorname{success}(\theta_{s,p,i,j},q,H)\}.
\end{equation}
Each task family contributes equally to the panel. For two selection rules, we report
\begin{equation}
\Delta_s(p,p')=\frac{1}{IJ}\sum_{i=1}^{I}\sum_{j=1}^{J}(R_{s,p,i,j}-R_{s,p',i,j}).
\end{equation}
We average sampling replicates within each initialization, then report pool-specific and equal-source means. All checkpoints are evaluated on the same task panel. The 16 initialization--sampling combinations represent eight distinct initializations.

\paragraph{Likelihood baselines.} Low, middle, and high NLL use canonical action-level scores from the fixed base student. Matched selectors choose within task/action strata and preserve the reference schedule's stratum at every position and its distinct-content quotas. Global high NLL instead ranks all distinct contents without those quotas. Global random samples 256 distinct contents uniformly without replacement; repeated original rows do not increase selection probability.

\paragraph{Optimizer-aware score.} At warmup checkpoint $c$, let $g_c(e)$ be a candidate's LoRA gradient, with frozen Adam moments $m_c,v_c$. The feature follows the LESS-style expression
\begin{equation}
a_c(e)=\frac{\beta_1 m_c+(1-\beta_1)g_c(e)}{\sqrt{\beta_2 v_c+(1-\beta_2)g_c(e)^2+\epsilon}}.
\end{equation}
This feature uses fixed optimizer states and omits AdamW bias correction. Let $u(x)=x/\|x\|_2$ and $b_{c,f}$ be the mean of individually normalized public-support gradients for family $f$. The score is
\begin{equation}
s(e)=\max_f\sum_{c\in\{8,16,24,32\}}\tfrac14\,u(a_c(e))^\top b_{c,f}.
\end{equation}
Checkpoint averaging precedes the maximum over support families. The 40 fixed public actions contain 18 placement and 22 cleaning actions; no query answers are used. Features are computed in FP32 with numerical checks. Student training starts from the original base; the warmup checkpoints are used only for scoring.

\paragraph{Common replay.} Each global rule selects 256 distinct message contents. Each set is replayed in three shuffled passes using the same schedule-generation procedure, yielding 768 presentations. The rules can select different contents and task/action distributions. This budget fixes presentations and optimizer updates. Token counts are reported separately; FLOPs are not matched.

\begin{algorithm}[t]
\caption{Global experience selection under the common training budget}
\label{alg:global}
\begin{algorithmic}[1]
\Require Fixed pool $D_s$, rule $p$, support $K=256$, replay seed $j$
\State Deduplicate exact message contents; retain the first representative index
\If{$p$ is global random}
  \State Draw $K$ representatives uniformly without replacement
\Else
  \State Rank representatives by canonical NLL or the fixed optimizer-aware score
  \State Select the top $K$, breaking score ties by representative index
\EndIf
\State Create three seeded shuffled passes, totaling 768 presentations
\For{each of the eight initialization/dropout seeds $i$}
  \State Cold-start the original base with a fresh LoRA adapter
  \State Run 192 optimizer updates; audit initial weights, consumed tokens, and schedule
  \State Evaluate with the shared public-example executor and frozen task panel
\EndFor
\State Retain both sources, every initialization, and all paired differences
\end{algorithmic}
\end{algorithm}

\section{Experimental Design}
\paragraph{Environment and data.} We use ALFWorld's text interface~\citep{shridhar2021alfworld}. The API and Generic pools contain 721 and 732 eligible action demonstrations, with 418 and 508 distinct message contents, from 35 and 29 collection tasks. We compare selectors within each historical pool. Because collection conditions were not randomized, the comparisons do not estimate the effect of the collection harness. Selection reuses the existing records without new teacher calls.

\paragraph{Student and training.} Qwen3-4B-Instruct-2507~\citep{yang2025qwen3,qwen2025instruct2507} is pinned to revision \texttt{cdbee75f17c01a7cc42f958dc650907174af0554}. LoRA~\citep{hu2022lora} uses rank 16, alpha 32, dropout 0.05, and $q/k/v/o$ attention projections. Every student receives 192 optimizer updates and 768 presentations, with global batch size four, learning rate $10^{-4}$, 0.05 warmup ratio, zero weight decay, and maximum sequence length 2,048. Paired selection rules share initial parameter hashes; processed input and target tokens are recorded for each student.

\paragraph{Shared executor.} Evaluation uses the same JSON action interface and two fixed public examples, task-local history length 12, environment seed 11, temperature zero, batch size eight, at most 40 calls per episode, and at most 128 generated tokens per call. There is no teacher card, retry, or cross-task memory. The inherited adapter interface combines each student's A adapter with an immutable zero-B branch; B is not updated and weights are not merged in BF16.

\paragraph{Experimental matrix.} The original seven-rule development matrix trains 224 students across two sources, eight initialization seeds, and two sampling seeds. Global random and high-NLL baselines add 64 students. Their comparison reuses 32 existing global-gradient students and the untrained base. All models use the same 128 fresh development groups, balanced between placement and cleaning. Groups are disjoint from training, interface-calibration, and reserved replication groups. All panels are drawn from the official training pool; the official seen/unseen tests are unused.

\paragraph{Analysis and reproducibility.} We report every planned condition, with mean success, initialization-level contrasts, and descriptive dispersion on the fixed panels. Sampling replicates are averaged within initialization. The analysis is descriptive: it reports no population p-values, confidence intervals, or equivalence tests. The 97 global-comparison endpoints are reconstructed from episode records and checked against training metadata and checkpoint identities. Appendix~\ref{app:reproducibility} describes the code and data records.

\section{Results}
\begin{table}[t]
\centering\small
\caption{\textbf{Complete development results.} Success is a percentage. Each source/rule mean averages eight initializations and two sampling seeds. Global rules have equal selection freedom. Matched rules share task/action quotas. The same-panel base is 67.188\%.}
\label{tab:main}
\begin{tabular}{llrr}\toprule
Space & Rule & API & Generic\\\midrule
Matched & Random & 77.344 & 68.652\\
& Low NLL & 72.070 & 67.627\\
& Middle NLL & 73.975 & 68.750\\
& High NLL & 78.906 & 68.408\\
& Gradient high & 77.295 & 71.875\\
& Gradient low & 72.559 & 65.186\\\midrule
Global & Random & 73.193 & 67.236\\
& High NLL & 77.344 & 69.141\\
& Gradient & \textbf{81.445} & \textbf{73.193}\\\bottomrule
\end{tabular}
\end{table}

\subsection{Equal Global Selection Freedom}
Global gradient selection reaches 81.445\% in API and 73.193\% in Generic (Table~\ref{tab:main}). Its gains over global random are +8.252 and +5.957 percentage points; gains over global high NLL are +4.102 and +4.053 points. All eight initialization-level mean differences are positive in each pool against both global comparators.

\begin{figure*}[t]\centering
\includegraphics[width=0.88\textwidth]{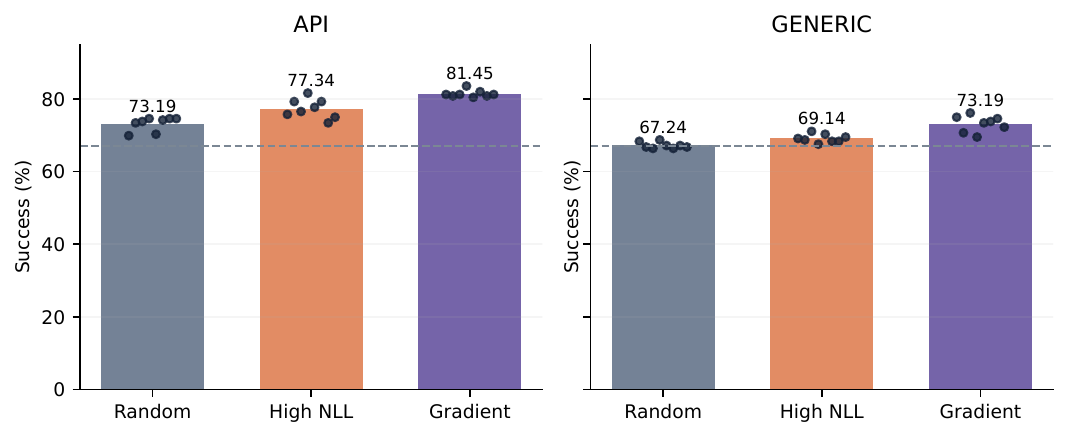}
\caption{\textbf{Global comparison on fresh development groups.} Each dot is an initialization's mean over two sampling seeds. Bars average eight initializations; the dashed line is the same-panel untrained base. The global baselines share support size and replay budget with the gradient rule.}
\label{fig:global}
\end{figure*}

Gradient selection retains an advantage when random and high-NLL selection have the same freedom to choose task and action coverage.

\subsection{Selection Criterion Interacts with Coverage Constraints}
\begin{figure*}[t]\centering
\includegraphics[width=0.88\textwidth]{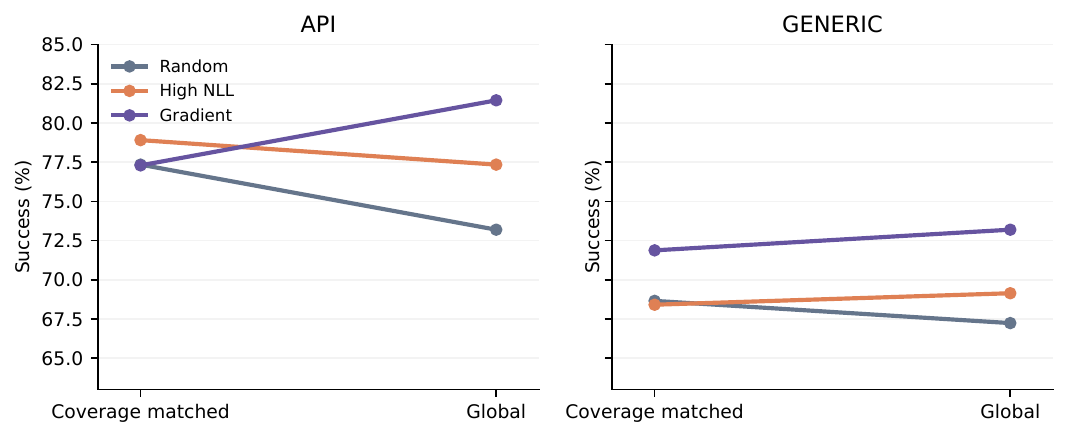}
\caption{\textbf{Selection rule $\times$ admissible coverage.} The common student and presentation budget are fixed. Relaxing the task/action quotas helps the gradient selector and hurts Random in both pools. Lines connect complete experimental means, not intermediate learning checkpoints.}
\label{fig:interaction}
\end{figure*}

Moving from matched to global selection changes Random by $-4.150$ points in API and $-1.416$ in Generic. The gradient rule changes by $+4.150$ and $+1.318$ points. High NLL changes by $-1.562$ and $+0.732$ points. The effect of selection freedom depends on the ranking rule (Figure~\ref{fig:interaction}).

Relaxing quotas changes content, frequency, and interaction history together. These comparisons therefore evaluate the complete selector and do not isolate the effect of coverage alone.

\subsection{Initialization-Level Consistency}
\begin{figure*}[t]\centering
\includegraphics[width=0.78\textwidth]{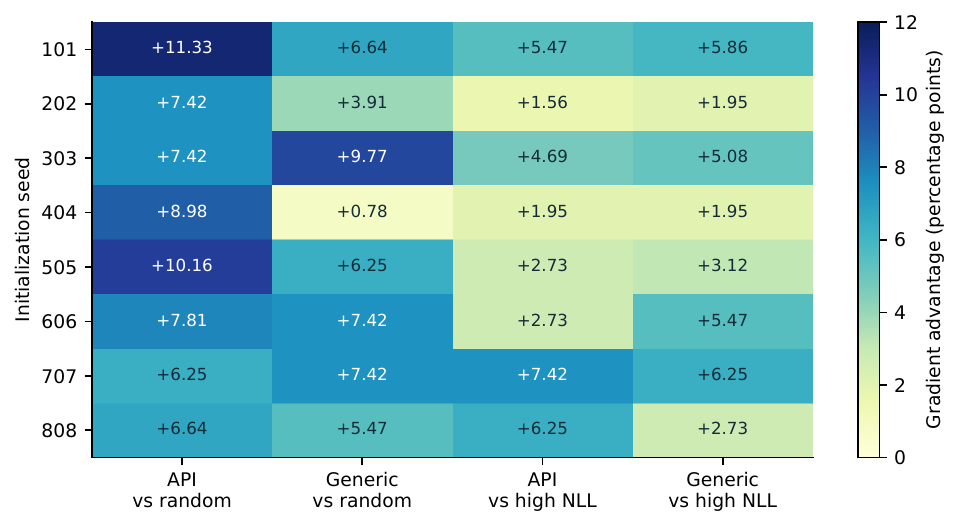}
\caption{\textbf{Paired initialization-level results.} Cells show gradient minus comparator success, in percentage points, after averaging the two sampling seeds. Each row corresponds to one of the eight initializations.}
\label{fig:seeds}
\end{figure*}

Figure~\ref{fig:seeds} shows every initialization-level contrast. Gains over global random range from 6.250 to 11.328 points in API and 0.781 to 9.766 in Generic. Gains over global high NLL range from 1.562 to 7.422 in API and 1.953 to 6.250 in Generic. The advantage holds for every initialization, with different magnitudes across the two pools. Each point averages two sampling replicates.

\subsection{Likelihood and Behavioral Utility}
Low, middle, and high NLL obtain 72.070\%, 73.975\%, and 78.906\% in the API matched conditions. The Generic results have smaller, nonmonotonic differences. Low-NLL selection underperforms the higher-NLL alternatives in API. High NLL remains a strong baseline: it exceeds matched gradient-high selection in API, even though global gradient selection leads both global baselines in each pool.

Appendix~\ref{app:scatter} plots all 36 source/rule/sampling schedules, each averaged over eight trained initializations. The schedules overlap, so the plot describes these comparisons rather than an independent corpus sample.

\section{Reserved-Task Replication}
\begin{figure*}[t]\centering
\includegraphics[width=0.9\textwidth]{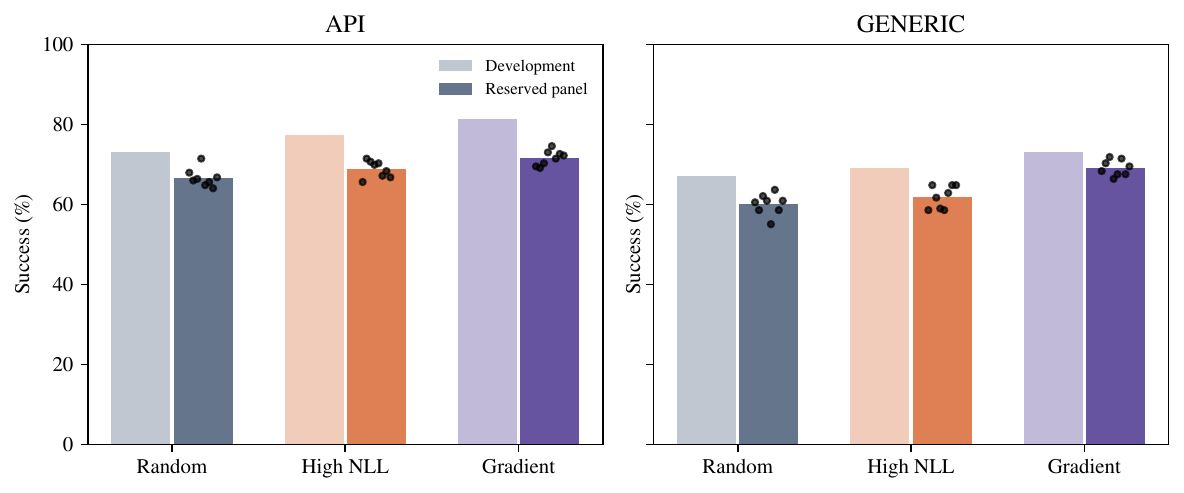}
\caption{Development and reserved-panel success for the same checkpoints. Light/dark bars show panel means; dots show reserved-panel results for each initialization, averaged over two sampling seeds.}
\label{fig:reserved}
\end{figure*}

The reserved evaluation uses 128 task groups disjoint from development, with all 97 checkpoints reused without further training. Its 12,416 episodes cover both experience pools, eight initializations, two sampling seeds, and the three global rules fixed before evaluation. Table~\ref{tab:reserved} reports success by rule and pool.
\begin{table}[H]\centering\small
\caption{Success on the reserved task panel (\%). Each trained-rule mean averages eight initializations and two sampling seeds.}
\label{tab:reserved}
\begin{tabular}{lrr}\toprule Rule & API & Generic\\\midrule
Global random & 66.650 & 60.059 \\
Global high NLL & 68.799 & 61.914 \\
Global gradient & 71.631 & 69.141 \\
\bottomrule\end{tabular}\end{table}
Relative to global random, the success differences are +4.980 points in API and +9.082 in Generic. Relative to global high NLL, the success differences are +2.832 points in API and +7.227 in Generic. After averaging sources and sampling replicates within each initialization, all eight contrasts are positive against both baselines. Figure~\ref{fig:reserved} compares the development and reserved-panel outcomes for the same checkpoints.

\section{Costs and Limitations}
\paragraph{Measured cost.} The 224-student development study used 62.146 cumulative GPU-hours, including preparation. Training and evaluating the 64 additional global baselines used 17.419 hours. The reserved-task evaluation reused 97 checkpoints and used 24.240 hours, for 41.659 hours across the two supplementary studies, within their 48-hour budget. The earlier source collection and historical experiments are accounted for separately.

\paragraph{Support and token budgets.} Gradient scoring uses 40 public target-family actions and requires feature warmup. These actions are not added to student SFT. Public examples are also available to the common deployment executor, but the selectors use different computations. Equal presentations do not imply equal FLOPs: API matched gradient-high inputs average 476,314 tokens versus 500,016 for gradient-low; Generic averages 510,404 versus 527,626. Presentation and update counts are matched; token and compute totals can differ.

\paragraph{Length and mechanism.} A CPU-only feasibility check fixes task/action budgets and target-token length while allowing input-length bins of 0, 8, 16, and 32 tokens. A usable control must keep input-token imbalance at most 1\% and retain at least 25\% of the original high/low gradient-score gap in every source/sampling condition. None passes. At width 32, API retains 3.45--5.23\% of the gap and Generic 17.06--18.17\%. The available pools cannot support this control while retaining the planned score contrast. Length and other content properties remain possible explanations for part of the selection effect.

\paragraph{Scope.} The evidence comes from one backbone, one text environment, and two fixed experience pools. The reserved panel tests new tasks within this setting. Long-term retention, repeated learning cycles, other backbones, and robot control remain untested. WebShop provides an interface calibration, not a second-environment selection result. Official ALFWorld seen/unseen benchmark sets were not used.

\section{Conclusion}
With the same global selection freedom and training budget, optimizer-aware gradient selection outperforms random and high-NLL selection in both experience pools. The gains persist on reserved tasks: 7.03 and 5.03 percentage points after averaging sources equally. Relaxing coverage constraints benefits the gradient rule but reduces random-selection performance. Evaluating a selection criterion therefore requires specifying the space from which it selects. This study tests one training stage; the next question is whether the selected experience also improves retention and repeated adaptation.

\begingroup\fontsize{8}{9}\selectfont\setlength{\bibsep}{0pt}
\bibliographystyle{plainnat}
\bibliography{references}
\endgroup

\appendix
\section{Recorded Training Dynamics}
\begin{figure}[H]\centering
\includegraphics[width=0.9\textwidth]{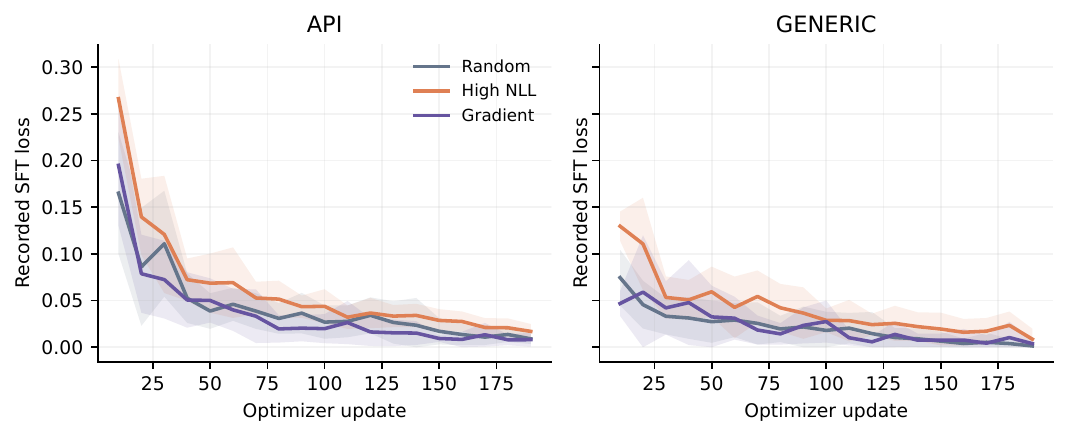}
\caption{Training loss at logging steps 10 through 190. Lines average 16 model histories per source/rule; shading shows their SD. Each student runs 192 updates.}
\end{figure}
Training loss falls rapidly during the first updates and then flattens. Loss was logged every ten updates, so the final two updates have no separate recorded point. Lower training loss does not consistently predict the ordering of task success.

\section{Aggregate Action Composition}
\begin{figure}[H]\centering
\includegraphics[width=0.92\textwidth]{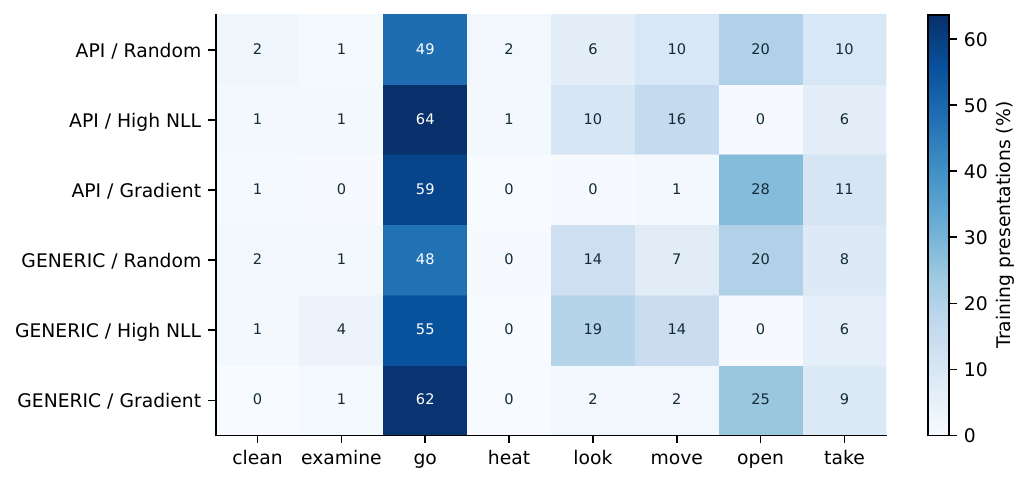}
\caption{Action-family composition of the selected 768-presentation schedules, averaged over the two sampling seeds. Values are percentages.}
\end{figure}
Global selectors can change these distributions. Matched selectors preserve the reference task/action positions and distinct-content quotas, allowing a separate comparison under fixed coverage.

\section{Imitation-Likelihood Scatter}\label{app:scatter}
\begin{figure}[H]\centering
\includegraphics[width=0.96\textwidth]{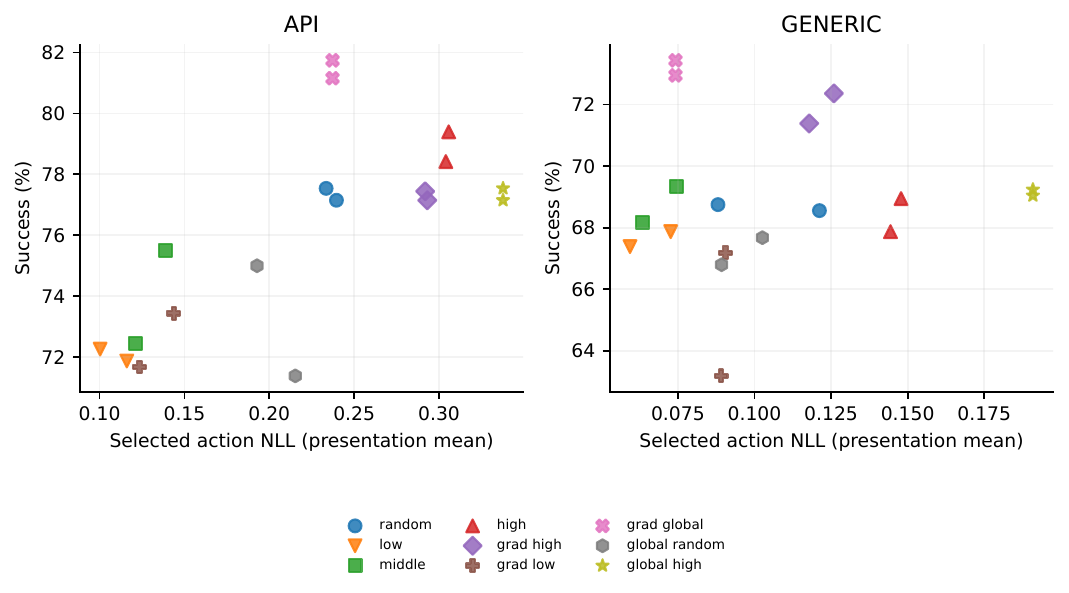}
\caption{All 36 source/rule/sampling schedules. Horizontal coordinates are mean selected action NLL weighted by training presentations; vertical coordinates average the eight student initializations. Schedules overlap, so the scatter is descriptive.}
\end{figure}

\section{Earlier Selection and Source Studies}
Earlier complete-pass scheduling improves success over subset replay by 5.176 and 1.270 points at the same update/presentation budget. Task-matched low-NLL selection is below Random by 17.773 and 7.812 points; task/action matching reduces the descriptive gap to 4.818 and 2.995 points. The separate canonical-scoring retraining comparison gives low-NLL-minus-Random differences of $-3.516$ and $-3.060$ points. These studies use older, repeatedly examined development panels and are reported separately from the fresh-panel comparisons.

In the earlier teacher-assisted versus ReAct comparison, the common-success matched adaptation-AUC difference is $-3.56$ points across three seeds. Within-pool selection addresses a different question from comparing collectors on all eligible experience.

\section{Reproducibility}\label{app:reproducibility}
The experiment records include protocols, scoring checks, training schedules, initialization records, aggregates, and figure data. The global-baseline study fixes 564 inputs, its parent development study fixes 509, and the reserved-panel evaluation fixes 1,025. Checkpoint identities and evaluation records are checked before aggregation. All 97 reserved-panel endpoints are included, and the three final cloud aggregates match the local files byte for byte. Raw tasks, conversations, requests, and model weights remain outside the manuscript bundle.

\clearpage
\section{Historical Matched-Selection Construction}
\label{sec:method}

\paragraph{Task panel.} The earlier studies use 32 cooling and 32 heating tasks, with four or six initialization seeds. The main study uses separate placement/cleaning panels and eight initializations. Their results are reported separately.
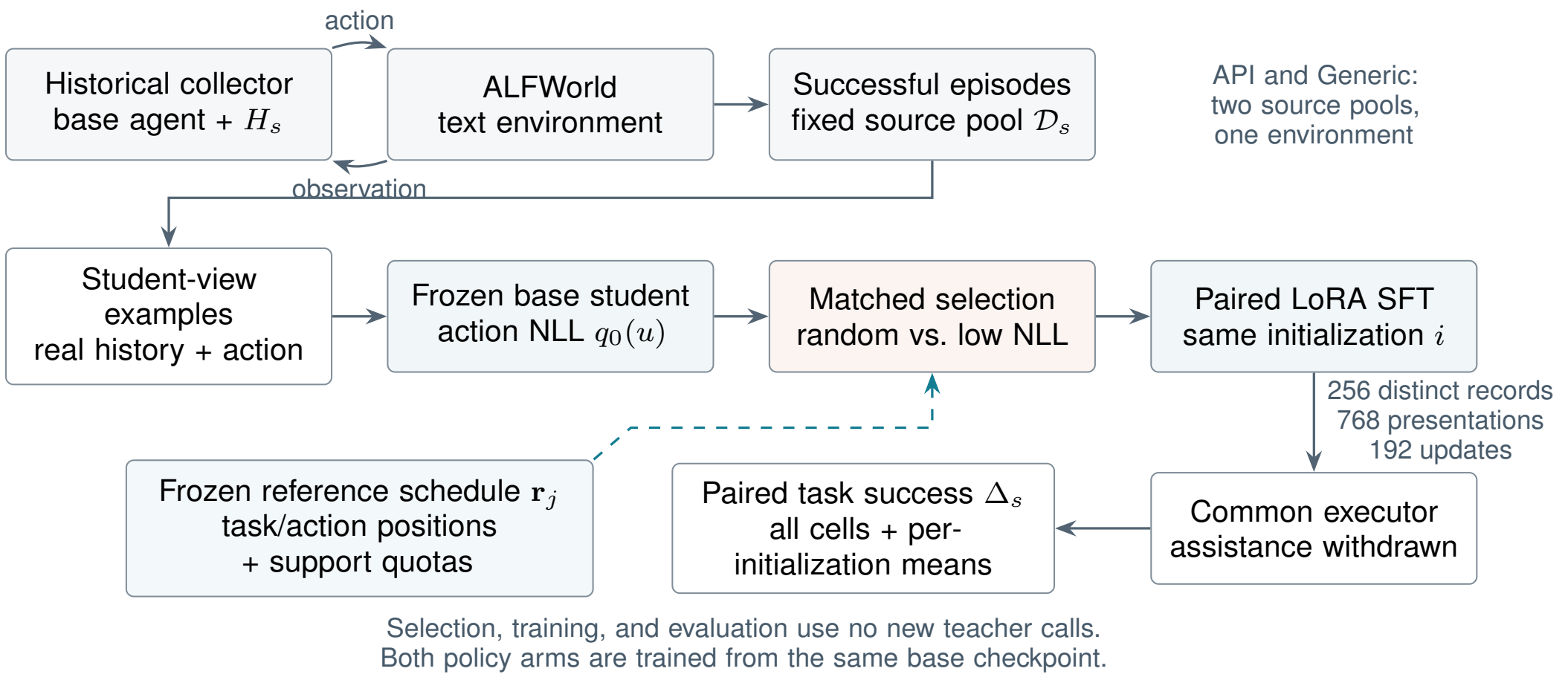
\begin{figure}[htbp]\centering
\resizebox{\linewidth}{!}{\begin{tikzpicture}[
  font=\sffamily\footnotesize,
  block/.style={draw=ink!55,rounded corners=2pt,align=center,text width=2.72cm,minimum height=1.05cm,inner sep=5pt,fill=white},
  flow/.style={-{Stealth[length=2mm]},line width=.65pt,draw=ink!75},
  control/.style={flow,dashed,draw=agentblue},
  note/.style={align=center,font=\sffamily\scriptsize,text=ink!80}]
\node[block,fill=ink!4] (harness) at (0,0) {Historical collector\\base agent + $H_s$};
\node[block,fill=ink!4] (env) at (3.6,0) {ALFWorld\\text environment};
\node[block,fill=ink!4] (pool) at (7.2,0) {Successful episodes\\fixed source pool $\D_s$};
\node[note,text width=2.7cm] at (10.8,0) {API and Generic:\\two source pools,\\one environment};
\draw[flow] (harness.north east) to[bend left=18] node[above,note]{action} (env.north west);
\draw[flow] (env.south west) to[bend left=18] node[below,note]{observation} (harness.south east);
\draw[flow] (env) -- (pool);
\node[block] (view) at (0,-2.0) {Student-view examples\\real history + action};
\node[block,fill=agentblue!6] (score) at (3.6,-2.0) {Frozen base student\\action NLL $q_0(u)$};
\node[block,fill=agentorange!7] (select) at (7.2,-2.0) {Matched selection\\random vs.\ low NLL};
\node[block,fill=agentblue!6] (train) at (10.8,-2.0) {Paired LoRA SFT\\same initialization $i$};
\draw[flow] (pool.south) -- ++(0,-.35) -| (view.north);
\draw[flow] (view) -- (score);
\draw[flow] (score) -- (select);
\draw[flow] (select) -- (train);
\node[block,text width=4.05cm,fill=agentblue!5] (quotas) at (1.8,-4.0) {Frozen reference schedule $\mathbf r_j$\\task/action positions + support quotas};
\node[block,text width=3.25cm] (result) at (6.55,-4.0) {Paired task success $\Delta_s$\\all cells + per-initialization means};
\node[block] (eval) at (10.8,-4.0) {Common executor\\assistance withdrawn};
\draw[control] (quotas.north east) -- ++(.35,.30) -| (select.south);
\draw[flow] (train) -- node[right,note]{256 distinct records\\768 presentations\\192 updates} (eval);
\draw[flow] (eval) -- (result);
\node[note,text width=12.8cm] at (5.4,-5.1) {Selection, training, and evaluation use no new teacher calls. Both policy arms are trained from the same base checkpoint.};
\end{tikzpicture}}
\caption{Earlier experimental pipeline: collection, student-view records, matched selection, paired training, and evaluation under a common executor. The main global-selection design is shown in Figure~\ref{fig:architecture}.}
\end{figure}

\subsection{Examples, predictability, and training schedules}
A fixed source pool $\D_s$ contains eligible action demonstrations $z=(x,h,a)$: task $x$, actual public interaction history $h$, and the next canonical action $a$. Here $a$ includes the supervised JSON serialization and its chat-template terminator; $c(a)$ parses the command's action family. Let $u(z)$ identify the complete student-view messages by exact content. Deduplication uses exact message equality. Selected records retain their original histories; they are not assembled into synthetic trajectories.

For the frozen base student $p_{\theta_0}$, action predictability is measured by token-average target NLL,
\begin{equation}
 \ell_0(z)=-\frac{1}{|a|}\sum_{v=1}^{|a|}
 \log p_{\theta_0}(a_v\mid h,a_{<v}).
 \label{eq:nll}
\end{equation}
The original cache can assign different numerical values to identical content in different batching layouts. The historical pre-canonical selection experiments used the recorded mean over copies of each content identity,
\begin{equation}
 q_0(u)=\frac{1}{|\{z\in\D_s:u(z)=u\}|}
 \sum_{z\in\D_s:u(z)=u}\ell_0^{\mathrm{cache}}(z).
 \label{eq:cache}
\end{equation}
This definition applies to the earlier cached-score experiments. The main comparisons use canonical scores; rescoring and paired retraining are reported in Appendix~\ref{sec:scoring}.

A sampling recipe $j$ defines a frozen reference schedule $\mathbf r_j=(r_1,\ldots,r_B)$ with $B=768$ presentations. For a stratum function $g$, define the number of presentations and distinct content identities assigned to stratum $g$:
\begin{equation}
 n_g=\sum_{k=1}^{B}\ind\{g(z_{r_k})=g\},\qquad
 K_g=\left|\{u(z_{r_k}):g(z_{r_k})=g\}\right|.
 \label{eq:quotas}
\end{equation}
The two policies select $K_g$ identities from the eligible distinct support $\U_g$. Random selection samples uniformly without replacement; low-NLL selection solves
\begin{equation}
 A_g^{\low}\in\underset{A\subseteq\U_g,\,|A|=K_g}{\arg\min}
 \sum_{u\in A}q_0(u).
 \label{eq:selection}
\end{equation}
Ties are resolved by the representative corpus index. Within each stratum, a fixed recipe-specific shuffle cycles through the selected identities, visiting every identity before repeating it. The final schedule interleaves these sequences in the reference order. For the resulting policy schedule $\boldsymbol\sigma^{p}_{s,j}$,
\begin{align}
 g(z_{\sigma_k^p})&=g(z_{r_k})\quad &&\text{for every presentation }k,\label{eq:positions}\\
 \left|\{u(z_{\sigma_k^p}):g(z_{\sigma_k^p})=g\}\right|&=K_g\quad &&\text{for every stratum }g.\label{eq:support}
\end{align}
The selected content may differ while these constraints hold exactly.

\subsection{Progressively stronger controls}
\begin{figure}[t]
\centering
\resizebox{\linewidth}{!}{\begin{tikzpicture}[font=\sffamily\footnotesize,
 card/.style={draw=ink!40,rounded corners=2pt,align=left,text width=3.85cm,minimum height=1.65cm,inner sep=7pt},
 flow/.style={-{Stealth[length=2mm]},draw=ink!65,line width=.65pt}]
\node[card,fill=ink!3] (s2) at (0,0) {\textbf{1. Fix training dose}\\Full pass vs.\ subset replay\\768 draws / 192 updates\\\textcolor{ink!65}{Task mixture can differ}};
\node[card,fill=agentblue!5] (s3) at (4.65,0) {\textbf{2. Match task exposure}\\Random vs.\ low NLL\\Same task at each position\\256 distinct records};
\node[card,fill=agentorange!6] (s4) at (9.3,0) {\textbf{3. Match task/action}\\Random vs.\ low NLL\\Same task/action positions\\Same stratum support};
\draw[flow] (s2) -- (s3);
\draw[flow] (s3) -- (s4);
\end{tikzpicture}}
\caption{\textbf{Progressive controls.} Each comparison fixes additional properties of the training schedule. Task/action matching also restricts eligible content and weakens the selection contrast, so differences across stages do not quantify a mediation effect.}
\label{fig:controls}
\end{figure}
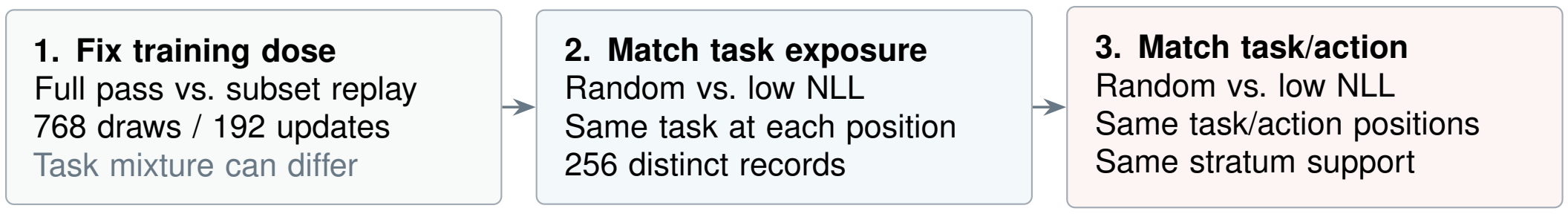
\smallpara{Training dose: complete pass versus subset replay.}
Both arms receive 768 presentations and 192 updates. Complete-pass scheduling traverses the eligible corpus before repeating rows; the replay arm reuses a 256-row subset. Membership, multiplicity, order, and task mixture can change together. This first comparison evaluates the scheduling intervention as a whole.

\smallpara{Task-matched selection.}
We set $g(z)=x$. The policies share task identity at every presentation, task counts, and 256 distinct task/content records. They can still differ in action composition. We cross four initialization/dropout seeds with four sampling recipes within each source.

\smallpara{Task-by-action-matched selection.}
We set $g(z)=(x,c(a))$ and inherit the task/action quotas from the task-matched random reference. The policies now share every task/action position and within-stratum support count. Action families absent from the reference remain absent. The support restriction retains approximately 66.15\% and 40.44\% of the preceding NLL-selection contrast in the API and Generic pools. The resulting comparison has a weaker score contrast as well as tighter coverage control. We cross six initialization/dropout seeds with the same four sampling recipes.

\begin{algorithm}[t]
\caption{Matched content selection and paired evaluation}
\label{alg:protocol}
\begin{algorithmic}[1]
\Require Fixed pool $\D_s$, base student $\theta_0$, reference $\mathbf r_j$, strata $g$
\State Compute content identities, scores $q_0(u)$, and quotas $(n_g,K_g)$
\For{$p\in\{\rand,\low\}$}
  \State Select $A_g^p$ independently within each reference stratum
  \State Cycle shuffled identities within each stratum; interleave in reference order
  \State Verify Eqs.~\eqref{eq:positions}--\eqref{eq:support}, total support, and schedule hashes
  \For{initialization/dropout seed $i$}
    \State Train a fresh LoRA adapter on $\boldsymbol\sigma^p_{s,j}$ for 192 updates
    \State Audit weights and evaluate through the common executor without collection guidance
  \EndFor
\EndFor
\State Report paired differences, all initialization means, and all wins/ties/losses
\end{algorithmic}
\end{algorithm}

\subsection{Training and estimands}
We fine-tune low-rank adapters~\citep{hu2022lora} while freezing the backbone. The supervised objective averages the token-normalized target loss over scheduled examples,
\begin{equation}
 \mathcal L(\phi;\boldsymbol\sigma)
 =\frac1B\sum_{k=1}^{B}\ell_{\theta_0+\Delta(\phi)}(z_{\sigma_k}).
 \label{eq:sft}
\end{equation}
The implementation uses microbatches of one example and gradient accumulation to a global batch of four. Both policies use the same initialization/dropout seed $i$, optimizer configuration, and update budget. Equal presentations do not imply identical input/target token totals or exact FLOPs.

Let $Y_{s,p,i,j,f,t}$ denote success for source $s$, policy $p$, initialization $i$, recipe $j$, family $f$, and task $t$. Each model's outcome and the primary contrast are
\begin{align}
 R_{s,p,i,j}&=\frac12\sum_{f\in\{\mathrm{cool},\mathrm{heat}\}}
                  \frac1{32}\sum_{t=1}^{32}Y_{s,p,i,j,f,t},\label{eq:success}\\
 \Delta_s&=\frac1{IJ}\sum_{i=1}^{I}\sum_{j=1}^{J}
                (R_{s,\low,i,j}-R_{s,\rand,i,j}).\label{eq:effect}
\end{align}
Initialization seeds and sampling recipes are crossed factors. We also report $\Delta_{s,i}=J^{-1}\sum_j(R_{s,\low,i,j}-R_{s,\rand,i,j})$. The 24 task/action-matched cells are not 24 independent initialization seeds. Conditional repeats are excluded from these primary means.

\clearpage
\section{Detailed Historical Selection Experiments}
\label{sec:experiments}
These studies use the earlier cooling/heating development panels. Their outcomes are reported separately from the main placement/cleaning comparisons.
\subsection{Setting and evaluation protocol}
We use Qwen3-4B-Instruct-2507 from the Qwen3 family~\citep{yang2025qwen3}, with a pinned revision recorded in Appendix~\ref{app:inventory}. The two historical experience pools, API and Generic, contain 721/732 eligible action rows from 35/29 collection tasks and 418/508 exact distinct message records. We compare selectors \emph{within} each historical source. Both pools come from the same environment; collection conditions were not randomized. The selection studies make no new teacher calls.

Training uses BF16 LoRA with rank 16, alpha 32, dropout 0.05, attention projections $q,k,v,o$, learning rate $10^{-4}$, zero weight decay, 0.05 warmup ratio, and maximum sequence length 2,048. Both policies receive 192 updates and 768 presentations. The task-matched comparison has 64 primary models, four conditional retrains, and two fresh full-pass reference checks. The task/action-matched comparison has 96 primary models and four conditional retrains. A paired policy cell shares a physical GPU; every source/policy group in the latter study has 12 primary models on each of two RTX 5090s.

Evaluation uses 32 cooling and 32 heating development tasks, environment seed 11, temperature zero, batch size eight, and a common JSON executor. Collection-specific assistance is removed; fixed public execution instructions and examples remain. The audited checkpoint combines the trained A adapter with an untrained zero-effect B adapter at step zero. These are repeatedly inspected development tasks from the official training pool. Official seen/unseen tests are unused. Results are paired descriptive comparisons.

\begin{table}[t]
\centering
\caption{\textbf{Completed selection comparisons.} Success rates are percentages; $\Delta$ is low NLL minus random in percentage points. W/T/L counts paired initialization--recipe cells, not independent initialization seeds. Conditional retrains are excluded.}
\label{tab:historical-main}
\small
\begin{tabular}{llrrrr}
\toprule
Control & Source & Random & Low NLL & $\Delta$ (pp) & W/T/L\\
\midrule
\SelectionRows
\bottomrule
\end{tabular}
\end{table}

\subsection{Experience support at a fixed training dose}
\label{sec:dose}
The complete-pass versus subset-replay study trains 68 models and evaluates 4,352 episodes. At 768 presentations and 192 updates, complete-pass exposure increases mean success by 5.176 points for API and 1.270 points for Generic. The source-contrast sampling standard deviations within the four initialization blocks change from 8.87/13.46/18.55/19.36 to 3.83/6.31/5.90/2.34 points. These values describe four recipes per block, not population variance intervals. Since the intervention can also change task mixtures and multiplicities, we next constrain task exposure directly.

\subsection{Low action NLL under matched task exposure}
\label{sec:task}
The task-matched study trains 70 models and evaluates 4,480 episodes. Its primary comparison uses four initialization seeds crossed with four recipes for each source and policy. Random and low-NLL schedules share task position, task counts, and 256 distinct task/content records. Low-NLL selection lowers mean success from 27.930\% to 10.156\% in API and from 12.695\% to 4.883\% in Generic (Table~\ref{tab:historical-main}). Every initialization-level average is negative in both pools.

\begin{figure}[t]
  \centering
  \includegraphics[width=\linewidth]{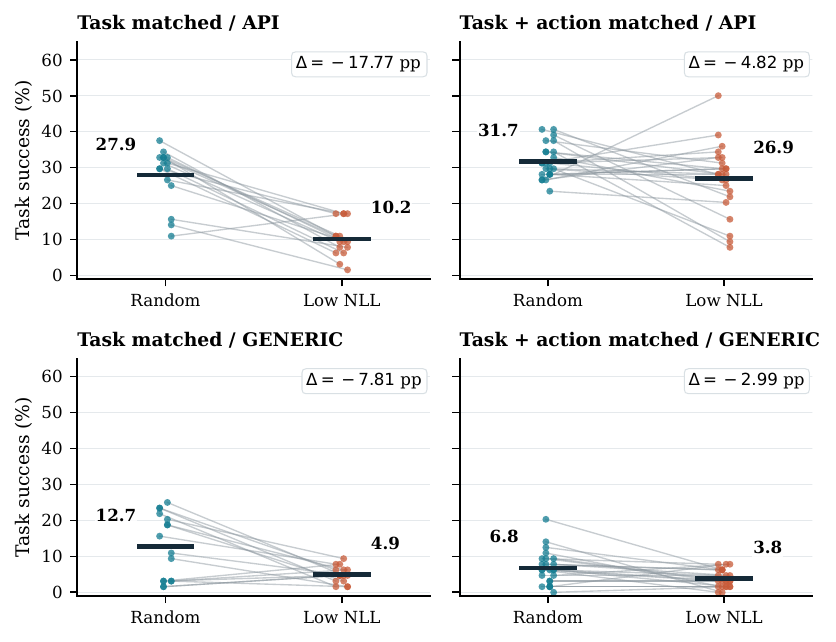}
  \caption{\textbf{Every primary paired cell.} Lines connect random and low-NLL outcomes for one initialization--recipe pair; black ticks show means. Top: API; bottom: Generic. Left: task matching (16 pairs per source). Right: task/action matching (24 pairs per source). All panels use the same vertical scale. No confidence interval is encoded. Lines sharing an initialization are not independent training seeds.}
  \label{fig:outcomes}
\end{figure}

Low NLL outperforms Random in one of 16 API cells and five of 16 Generic cells (Figure~\ref{fig:outcomes}). Action composition remains unconstrained in this comparison, so these results do not isolate content properties beyond action coverage.

\subsection{Residual differences after action-composition matching}
\label{sec:action}
The task/action-matched study trains 100 models and evaluates 6,400 episodes, using 14.663 cumulative GPU-hours against an 18-hour cap. The 96 primary models cover six initialization/dropout seeds and four recipes. Random and low NLL achieve 31.706\% and 26.888\% for API, and 6.771\% and 3.776\% for Generic, respectively. API has five negative initialization-level means and one zero; Generic has six negative means. Paired W/T/L counts are 8/1/15 and 3/6/15, respectively.

The gap narrows after task/action matching, although this comparison cannot identify the contribution of action composition alone. The reference quotas change available content support and weaken the selection contrast. Absolute performance also changes differently across pools. A post-hoc comparison restricted to the first two initialization blocks changes API random/low-NLL success from 28.125\%/10.938\% to 32.031\%/24.023\%; Generic changes from 11.133\%/5.469\% to 6.250\%/3.906\%. Gap reduction can accompany improvement or deterioration in absolute success. Appendix~\ref{app:seeds} reports all initialization-level differences.

\subsection{Training variability and scoring consistency}
\label{sec:scoring}
Four prespecified conditional retrains in the final study use initialization 101, recipe 101, and the original physical GPUs. All four produce different final weights. Success changes are 0.000 and $-4.688$ points for API random and low NLL, and $+1.562$ and $-1.562$ for Generic. These conditional repeats are reported separately from the primary means. An earlier post-hoc check reevaluates two stored checkpoints under their original configurations, adding 128 episodes; each reproduces its own recorded success and step count. This reproducibility check is limited to the two stored checkpoints.

An independent scoring diagnostic completes 1,194 forward passes, including uniform scoring of all 853 distinct message inputs, in 107.127 GPU seconds. Canonical repeats and original-layout replays reproduce their respective scores exactly. Four checks against direct supervised loss have absolute discrepancy at most $3.54\times10^{-6}$. Padding/precision stress tests nevertheless reveal numerical differences. The 32 stress inputs were selected deliberately and are not a representative error sample.

Uniform scores change 4--11 of 256 selected records across 16 original recipes, retaining 95.703\%--98.438\% of the content. Paired retraining of 96 models gives low-NLL-minus-Random differences of $-3.516$ points in API and $-3.060$ in Generic. These are separate retraining results; the earlier cached-score comparisons remain in Table~\ref{tab:historical-main}. Appendix~\ref{app:scoring} reports the numerical checks.

\subsection{Source comparison and second-environment calibration}
\label{sec:scope}
The teacher-assisted versus ReAct self-collection study compares entire collection pipelines. Collection succeeds in 69/144 and 110/144 attempts, respectively. On 58 common successful task/seed pairs, the API-minus-ReAct raw adaptation-AUC difference is $-3.56$ points, with seed values $-9.77$, $-3.42$, and $+2.49$. Common-success matching removes unmatched coverage. A full-eligible-data training comparison remains open, and the within-pool selection results do not answer that source-comparison question.

A WebShop base-model calibration completes 64 train-only episodes on 32 shared goals under two public prompt ports. ReAct solves 5/32 and Act-only solves 6/32; their mean partial scores are 0.4125 and 0.4380. All 463 requests finish without transport errors, but include 34 invalid-format and 136 invalid-control model decisions. The pilot measures base-model performance and interface errors before a future learning comparison. It includes no SFT intervention. Full details appear in Appendix~\ref{app:webshop}.

\clearpage
\section{Experimental Inventory and Implementation}
\label{app:inventory}
Table~\ref{tab:inventory} separates new training from checkpoint reuse and diagnostic evaluations. The first three selection studies train 238 models and run 15,232 episodes. Subsequent studies add canonical-score retraining and the placement/cleaning comparisons. The reserved-panel study evaluates existing checkpoints. The scoring diagnostic performs forward passes only, and WebShop calibration adds 64 episodes without training. Reused tasks and checkpoints are identified separately in the inventory.

\begin{table}[h]
\centering
\caption{Completed study inventory. Additional references and conditional repeats are excluded from primary selection means.}
\label{tab:inventory}
\small
\resizebox{\linewidth}{!}{\begin{tabular}{lrrl}
\toprule
Study & New trained models & New episodes & Primary design / additional checks\\
\midrule
Complete pass vs.\ replay & 68 & 4,352 & 64 primary + 4 repeats\\
Task matching & 70 & 4,480 & 64 primary + 4 repeats + 2 references\\
Task/action matching & 100 & 6,400 & 96 primary + 4 repeats\\
Canonical-score replication & 96 & 6,144 & 6 initializations $\times$ 4 recipes, 2 sources\\
Fresh seven-rule development & 224 & 28,800 & 224 students plus an untrained base\\
Global missing baselines & 64 & 8,192 & 2 global rules; 32 gradient students reused\\
Reserved-task evaluation & 0 & 12,416 & 97 existing checkpoints, no new training\\
Stored-checkpoint reevaluation & 0 & 128 & 2 checkpoints, original configurations\\
Uniform scoring diagnostic & 0 & 0 & 1,194 forwards; 853 distinct inputs\\
WebShop base calibration & 0 & 64 & 32 shared goals, two public prompts\\
\bottomrule
\end{tabular}}
\end{table}

\begin{table}[h]
\centering
\caption{Shared configuration for the controlled selection studies.}
\label{tab:config}
\small
\begin{tabular}{ll}
\toprule
Parameter & Value\\
\midrule
Backbone & Qwen3-4B-Instruct-2507, fixed revision\\
Revision & \texttt{cdbee75f17c01a7cc42f958dc650907174af0554}\\
Training precision / adaptation & BF16 / LoRA\\
LoRA rank / alpha / dropout & 16 / 32 / 0.05\\
Trainable projections & \texttt{q\_proj}, \texttt{k\_proj}, \texttt{v\_proj}, \texttt{o\_proj}\\
Learning rate / weight decay & $10^{-4}$ / 0\\
Warmup ratio / maximum sequence & 0.05 / 2,048 tokens\\
Microbatch / gradient accumulation & 1 example / 4 microbatches\\
Updates / sample presentations & 192 / 768\\
Distinct task/content budget (selection) & 256 per model\\
Historical evaluation & 32 cooling + 32 heating tasks, equal weights\\
Fresh development / reserved panels & 64 placement + 64 cleaning groups per panel\\
Environment seed / temperature / batch & 11 / 0 / 8\\
Collection-specific guidance & Removed during evaluation\\
Official final evaluation sets & Unopened\\
\bottomrule
\end{tabular}
\end{table}

\smallpara{Checkpoint composition.}
Evaluation serves the trained A adapter together with an untrained B adapter at optimizer step zero. B has zero effect at initialization but remains part of the serialized composition. Audits check the base revision, A weights, zero-B identity, and full composition revision. In Eq.~\eqref{eq:sft}, $\Delta(\phi)$ denotes the trainable A-adapter update.

\smallpara{Consumption and budget accounting.}
Presentations, processed input tokens, supervised target tokens, initialized parameters, trained weights, schedules, and physical GPU assignments are recorded separately. Time accounting includes model loading, training, serving, diagnostics, interrupted attempts, and shutdown. The complete task/action study uses 14.663 cumulative GPU-hours with a fixed 18-hour cap. No direction-dependent early stop, seed replacement, or queue extension is used.

\clearpage
\section{Initialization-Level Results and Conditional Repeats}
\label{app:seeds}
Table~\ref{tab:seeds} reports initialization-level means over four sampling recipes for the task/action comparison. API initialization 404 has zero mean difference, with both positive and negative cells. The conditional repeats in Table~\ref{tab:repeats} keep the original initialization, recipe, and physical GPU, but produce different final trained parameters.

\begin{table}[h]
\centering
\caption{Task/action matching: low NLL minus random, in percentage points.}
\label{tab:seeds}
\begin{tabular}{rrr}
\toprule
Initialization/dropout seed & API & Generic\\
\midrule
\InitializationRows
\bottomrule
\end{tabular}
\end{table}

\begin{figure}[h]
\centering
\includegraphics[width=\linewidth]{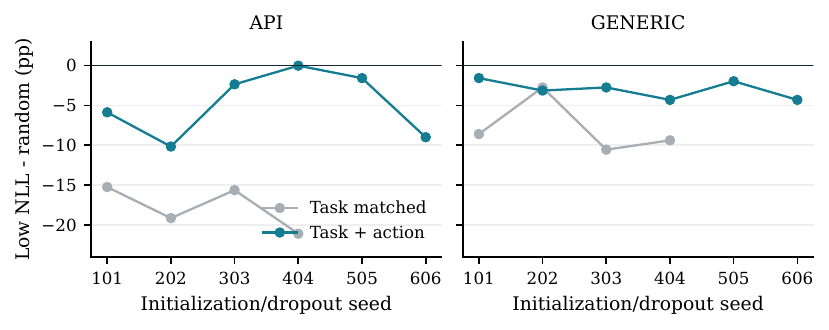}
\caption{\textbf{Initialization-level effects.} Each point averages four sampling recipes. Task matching has four initialization levels; task/action matching has six. Points summarize the crossed-seed comparisons without confidence intervals. Differences between curves also reflect changed content support and selection strength.}
\label{fig:initialization}
\end{figure}

\begin{table}[h]
\centering
\caption{Four prespecified task/action-study conditional repeats. Initialization 101 and recipe 101 throughout. Success rates are percentages.}
\label{tab:repeats}
\begin{tabular}{llrrr}
\toprule
Source & Policy & Original & Repeat & Change (pp)\\
\midrule
API & Random & 31.250 & 31.250 & 0.000\\
API & Low NLL & 34.375 & 29.688 & $-4.688$\\
Generic & Random & 4.688 & 6.250 & $+1.562$\\
Generic & Low NLL & 4.688 & 3.125 & $-1.562$\\
\bottomrule
\end{tabular}
\end{table}

\clearpage
\section{Scoring Geometry and Selection Sensitivity}
\label{app:scoring}
The completed diagnostic freezes 32 stress inputs: 16 selected for large recorded cache spans and 16 selected by content hash from the remainder. It evaluates canonical unpadded inputs with explicit positions and a fixed logits window, a canonical repeat, unpadded default positions, a dynamic logits window, and four padded conditions. It separately evaluates FP32 references, four direct supervised-loss checks, 392 original-layout replay observations, and uniform BF16 scores for all 853 distinct message inputs. The diagnostic comprises 1,194 forward passes, with no training updates, generation, or teacher calls.

Canonical repeat, unpadded default-position, and dynamic-window conditions match canonical scores exactly on the 32 stress inputs. Across padded conditions, mean absolute score deviations range from 0.0362 to 0.0603 nat/token. BF16 versus FP32 deviation has mean 0.0396 and maximum 0.1630, with rank correlation 0.9857 on the selected inputs. Original batch layouts reproduce their original cached values exactly. These numerical differences are measured on the prespecified stress set.

\begin{figure}[htbp]
\centering
\includegraphics[width=\linewidth]{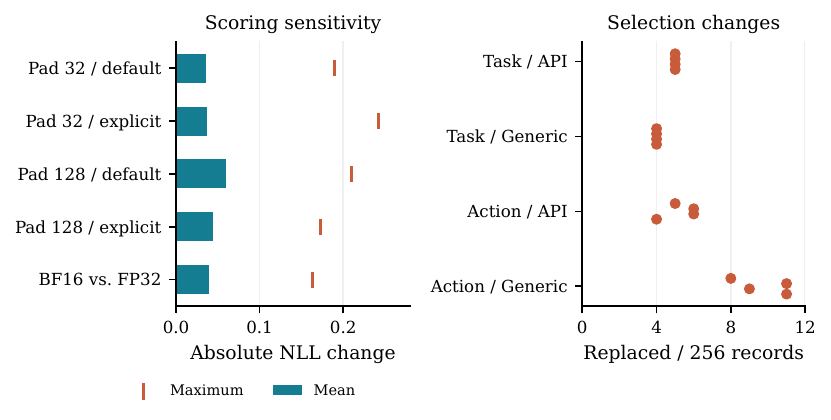}
\caption{\textbf{Completed scoring diagnostic.} Left: numerical layout sensitivity on the deliberately selected stress set. Right: content replacement after complete uniform rescoring, with each point a frozen selection recipe. Paired retraining using the canonical scores is reported below.}
\label{fig:scoring}
\end{figure}

\begin{table}[h]
\centering
\caption{Content records replaced out of 256 after uniform rescoring. Columns are sampling-recipe seeds. Original recipes are regenerated exactly before replacement is counted.}
\label{tab:replacement}
\begin{tabular}{llrrrr}
\toprule
Control & Source & 101 & 202 & 303 & 404\\
\midrule
Task matched & API & 5 & 5 & 5 & 5\\
Task matched & Generic & 4 & 4 & 4 & 4\\
Task/action matched & API & 5 & 6 & 6 & 4\\
Task/action matched & Generic & 8 & 11 & 9 & 11\\
\bottomrule
\end{tabular}
\end{table}

\smallpara{Completed canonical-score replication.}
The canonical-score study retrains 96 students: six initialization/dropout seeds, four recipes, two sources, and two policies, using 192 updates and the same 64 earlier tasks. Random indices are unchanged; low-NLL schedules use uniform scores. The study used 14.350 cumulative GPU-hours with zero failed models. API Random/Low NLL success is 29.688\%/26.172\%, a $-3.516$ point difference; Generic is 6.836\%/3.776\%, a $-3.060$ point difference. API has six negative initialization means; Generic has five negative and one positive. These results use the earlier panel, independently of the fresh-task comparisons.

\clearpage
\section{Second-Environment Base Calibration}
\label{app:webshop}
The WebShop pilot uses the full product/search backend and the pinned base checkpoint without an adapter. It evaluates 32 shared training goals under public ReAct and Act-only prompt ports, with at most 14 model decisions per episode and at most 100 output tokens per decision. All decisions, including thoughts and invalid controls, consume the episode budget. The backend is not queried through the ALFWorld training model servers.

\begin{table}[h]
\centering
\caption{Train-only WebShop calibration. Model-output errors are distinct from transport errors; all requests complete. This is not a post-SFT selection comparison.}
\label{tab:webshop}
\small
\begin{tabular}{lrr}
\toprule
Metric & ReAct & Act-only\\
\midrule
Goals solved / evaluated & 5/32 & 6/32\\
Success (\%) & 15.625 & 18.750\\
Mean partial score & 0.4125 & 0.4380\\
Model requests & 247 & 216\\
Physical actions & 86 & 91\\
Thought decisions & 78 & 38\\
Invalid format / invalid control & 24 / 59 & 10 / 77\\
Episodes exhausting decision budget & 10 & 10\\
Transport errors & 0 & 0\\
\bottomrule
\end{tabular}
\end{table}

\begin{figure}[htbp]
\centering
\includegraphics[width=\linewidth]{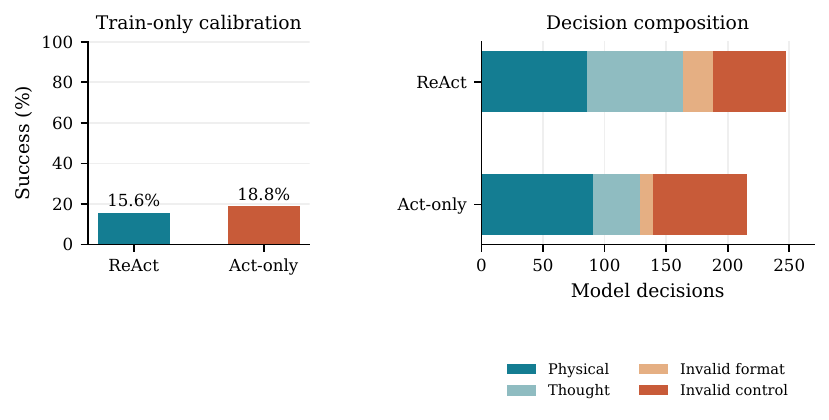}
\caption{\textbf{WebShop starting-point calibration.} Two public prompt ports on the same 32 training goals. The prompts differ by one solved goal. Decision counts show substantial invalid-format and invalid-control behavior.}
\label{fig:webshop}
\end{figure}

Across both prompts the pilot records 703,520 input and 15,844 output tokens, 463 completed requests, and 243.564 GPU seconds. Teacher calls, official evaluation episodes, and optimizer updates are zero. A second-environment \emph{learning} comparison must still fix the candidate pool, selection unit, training budget, common executor, and fresh task protocol.

\clearpage
\section{Source Comparison and Selector Scope}
\label{app:baseline}
The API-versus-ReAct source study is distinct from the API/Generic pools used in the controlled selection experiments. Both source collectors make 144 attempts using the same base student. The API condition uses cached guidance from public resets; ReAct uses public examples and reasoning/action interaction. The conditions also differ in format and thought mechanism, so that comparison is a pipeline comparison, not an isolated teacher-card intervention.

Common-success matching uses 58 paired successful task/environment-seed cases and yields 452 eligible action labels per source. At 256 A-training presentations, the raw adaptation-AUC difference is $-3.564$ points. The full eligible corpora instead contain 726 API and 981 ReAct labels. A prepared 1,024-presentation schedule visits each complete corpus at least once, but its formal source-comparison study has not been executed. The source-comparison result is reported separately from the within-pool selection analysis.

\smallpara{Scope relative to existing selectors.}
The low-NLL comparator selects action records and uses ordinary supervised target loss. SmartAD selects complete correct teacher trajectories and weights their segments; Rho-1 uses a reference-model token criterion. The gradient comparator implements a LESS-style optimizer-aware score, with its public support and preparation cost included in the study accounting. Published benchmark scores from different task splits are not directly comparable to these experiments.

\smallpara{Selection units and retained complete trajectories.}
Each action record includes its original preceding history. The source audit traces 117 successful trajectories, but only two API tasks and one Generic task have multiple distinct, fully retained eligible trajectories. A complete-trajectory selection comparison would require additional candidate trajectories. Regrouping the existing records leaves this candidate shortage unchanged.

\smallpara{Provenance and reproducibility.}
The manuscript bundle contains \LaTeX{} source, editable diagrams, vector figures, and the bibliography. Figure/table code and aggregate evidence are maintained with the project. Raw goals, messages, scoring journals, credentials, and adapter weights are excluded from the manuscript package. The NeurIPS style file is preserved with its recorded source and checksum; the document uses preprint mode.

\smallpara{Experimental setting.}
The experiments use simulated ALFWorld and WebShop tasks. They involve no physical robot actions, purchases, or new human-subject experiments.

\end{document}